\pdfoutput=1

\def\year{2020}\relax
\documentclass[letterpaper]{article} 
\usepackage{aaai20}  
\usepackage{times}  
\usepackage{helvet} 
\usepackage{courier}  
\usepackage[hyphens]{url}  
\usepackage{graphicx} 
\usepackage{graphicx}  
\usepackage{multirow}
\usepackage{booktabs}
\usepackage{bm}
\usepackage{amsmath}
\usepackage{array}
\usepackage{amssymb}

\nocopyright
\title{Region Normalization for Image Inpainting}
\author{Tao Yu,
	Zongyu Guo,
	Xin Jin,
	Shilin Wu,
	Zhibo Chen\thanks{Corresponding author},
	Weiping Li,
	Zhizheng Zhang,
	Sen Liu\\ 
CAS Key Laboratory of Technology in Geo-spatial Information Processing and Application System,\\
University of Science and Technology of China\\
\{yutao666, guozy, jinxustc, shilinwu\}@mail.ustc.edu.cn, \{chenzhibo, wpli\}@ustc.edu.cn,\\ zhizheng@mail.ustc.edu.cn, elsen@iat.ustc.edu.cn 
}
 
\begin{document}

\maketitle

\begin{abstract}
Feature Normalization (FN) is an important technique to help neural network training, which typically normalizes features across spatial dimensions. Most previous image inpainting methods apply FN in their networks without considering the impact of the corrupted regions of the input image on normalization, $e.g.$ mean and variance shifts. In this work, we show that the mean and variance shifts caused by full-spatial FN limit the image inpainting network training and we propose a spatial region-wise normalization named Region Normalization (RN) to overcome the limitation. RN divides spatial pixels into different regions according to the input mask, and computes the mean and variance in each region for normalization. We develop two kinds of RN for our image inpainting network: (1) Basic RN (RN-B), which normalizes pixels from the corrupted and uncorrupted regions separately based on the original inpainting mask to solve the mean and variance shift problem; (2) Learnable RN (RN-L), which automatically detects potentially corrupted and uncorrupted regions for separate normalization, and performs global affine transformation to enhance their fusion. We apply RN-B in the early layers and RN-L in the latter layers of the network respectively. Experiments show that our method outperforms current state-of-the-art methods quantitatively and qualitatively. We further generalize RN to other inpainting networks and achieve consistent performance improvements. Our code is available at https://github.com/geekyutao/RN.
\end{abstract}

\section{Introduction}

Image inpainting aims to reconstruct the corrupted (or missing) regions of the input image. It has many applications in image editing such as object removal, face editing and image disocclusion. A key issue in image inpainting is to generate visually plausible content in the corrupted regions.

Existing image inpainting methods can be divided into two groups: traditional and learning-based methods. The traditional methods fill the corrupted regions by diffusion-based methods \cite{bertalmio2000image,ballester2001filling,esedoglu2002digital,bertalmio2003simultaneous} that propagate neighboring information into them, or patch-based methods \cite{drori2003fragment,barnes2009patchmatch,xu2010image,darabi2012image} that copy similar patches into them. The learning-based methods commonly train neural networks to synthesize content in the corrupted regions, which yield promising results and have significantly surpassed the traditional methods in recent years. 
\begin{figure}[t]
	\begin{center}
		\includegraphics[width=.95\columnwidth]{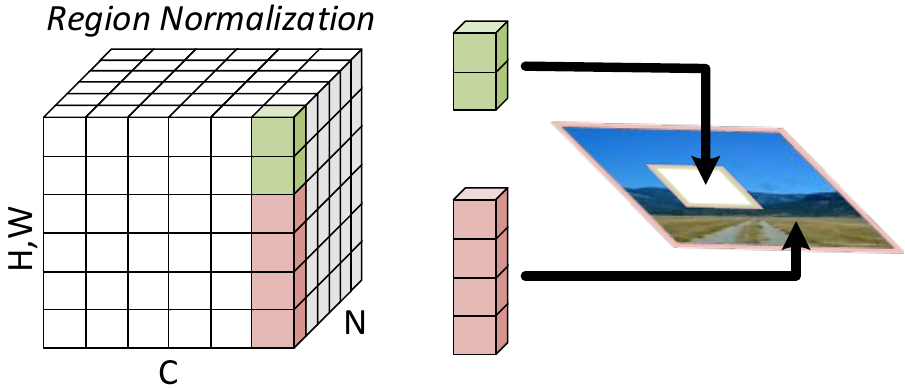}
	\end{center}
	\caption{Illustration of our Region Normalization (RN) with region number $K=2$. Pixels in the same color (green or pink) are normalized by the same mean and variance. The corrupted and uncorrupted regions of the input image are normalized by different means and variances.}
	\label{rn}
\end{figure}
Recent image inpainting works, such as \cite{yu2018generative,liu2018image,yu2019free,nazeri2019edgeconnect}, focus on the learning-based methods. Most of them design an advanced network to improve the performance, but ignore the inherent nature of image inpainting problem: unlike the input image of general vision task, the image inpainting input image has corrupted regions that are typically independent of the uncorrupted regions. Inputing a corrupted image as a general spatially consistent image into a neural network has potential problems, such as convolution of invalid (corrupted) pixels and mean and variance shifts of normalization. Partial convolution \cite{liu2018image} is proposed to solve the invalid convolution problem by operating on only valid pixels, and achieves a performance boost. However, none of existing methods solve the mean and variance shift problem of normalization in inpainting networks. In particular, most existing methods apply feature normalization (FN) in their networks to help training, and existing FN methods typically normalize features across spatial dimensions, ignoring the corrupted regions and resulting in mean and variance shifts of normalization. 

In this work, we show in theory and experiment that the mean and variance shifts caused by existing full-spatial normalization limit the image inpainting network training. To overcome the limitation, we propose Region Normalization (RN), a spatially region-wise normalization method that divides spatial pixels into different regions according to the input mask and computes the mean and variance in each region for normalization. RN can effectively solve the mean and variance shift problem and improve the inpainting network training. 

We further design two kinds of RN for our image inpainting network: Basic RN (RN-B) and Learnable RN (RN-L). In the early layers of the network, the input image has large corrupted regions, which results in severe mean and variance shifts. Thus we apply RN-B to solve the problem by normalizing corrupted and uncorrupted regions separately. The input mask of RN-B is obtained from the original inpainting mask. After passing through several convolutional layers, the corrupted regions are fused gradually, making it difficult to obtain a region mask from the original mask. Therefore, we apply RN-L in the latter layers of the network, which learns to detect potentially corrupted regions by utilizing the spatial relationship of the input feature and generates a region mask for RN. Additionally, RN-L can also enhance the fusion of corrupted and uncorrupted regions by global affine transformation. RN-L not only solves the mean and variance shift problem, but also boosts the reconstruction of corrupted regions. 

We conduct experiments on Places2 \cite{zhou2017places} and CelebA \cite{liu2015deep} datasets. The experimental results show that, with the help of RN, a simple backbone can surpass current state-of-the-art image inpainting methods. In addition, we generalize our RN to other inpainting networks and yield consistent performance improvements.

Our contributions in this work include:
\begin{quote}
	\begin{itemize}
		\item Both theoretically and experimentally, we show that existing full-spatial normalization methods are sub-optimal for image inpainting.
		\item To the best our knowledge, we are the first to \mbox{propose} spatially region-wise normalization $i.e.$ \mbox{Region Normalization} (RN).
		\item We propose two kinds of RN for image inpainting and the use of them for achieving state-of-the-art on image inpainting. 
	\end{itemize}
\end{quote}

\section{Related Work}
\subsection{Image Inpainting}
Previous works in image inpainting can be divided into two categories: traditional and learning-based methods. 

Traditional methods use diffusion-based \cite{bertalmio2000image,ballester2001filling,esedoglu2002digital,bertalmio2003simultaneous} or patch-based \cite{drori2003fragment,barnes2009patchmatch,xu2010image,darabi2012image} methods to fill the holes. The former propagate neighboring information into holes. The latter typically copy similar patches into the holes. The performance of these traditional methods is limited since they cannot use semantic information.

Learning-based methods can learn to extract semantic information by massive data training, and thus significantly improve the inpainting results. These methods map a corrupted image directly to the completed image. ContextEncoder \cite{pathak2016context}, one of pioneer learning-based methods, trains a convolutional neural network to complete image. With the introduction of generative adversarial networks (GANs) \cite{goodfellow2014generative}, GAN-based methods \cite{yeh2017semantic,iizuka2017globally,yu2018generative,xiong2019foreground,nazeri2019edgeconnect} are widely used in image inpainting. ContextualAttention \cite{yu2018generative} is a popular model with coarse-to-fine architecture. Considering that there are valid/uncorrupted and invalid/corrupted regions in a corrupted image, partial convolution \cite{liu2018image} operates on only valid pixels and achieves promising results. Gated convolution \cite{yu2019free} generalizes PConv by a soft distinction of valid and invalid regions. EdgeConnect \cite{nazeri2019edgeconnect} first predicts the edges of the corrupted regions, then generates the completed image with the help of the predicted edges. 

However, most existing inpainting methods ignore the impact of corrupted regions of the input image on normalization which is a crucial technique for network training.

\subsection{Normalization}
Feature normalization layer has been widely applied in deep neural networks to help network training.

Batch Normalization (BN) \cite{ioffe2015batch}, normalizing activations across batch and spatial dimensions, has been widely used in discriminative networks for speeding up convergence and improve model robustness, and found also effective in generative networks. Instance Normalization (IN) \cite{ulyanov2016instance}, distinguished from BN by normalizing activations across only spatial dimensions, achieves a significant improvement in many generative tasks such as style transformation. Layer Normalization (LN) \cite{ba2016layer} normalizes activations across channel and spatial dimensions ($i.e.$ normalizes all features of an instance), which helps recurrent neural network training. Group Normalization (GN) \cite{wu2018group} normalizes features of grouped channels of an instance and improves the performance of some vision tasks such as object detection. 

Different from a single set of affine parameters in the above normalization methods, conditional normalization methods typically use external data to reason multiple sets of affine parameters.
Conditional instance normalization (CIN) \cite{dumoulin2016learned}, adaptive instance normalization (AdaIN) \cite{huang2017arbitrary}, conditional batch normalization (CBN) \cite{de2017modulating} and spatially adaptive denormalization (SPADE) \cite{park2019semantic} have been proposed in some image synthesis tasks.

None of existing normalization methods considers spatial distribution's impact on normalization.

\section{Approach}
In this secetion, we show that existing full-spatial normalization methods are sub-optimal for image inpianting problem as motivation for Region Normalization (RN). We then introduce two kinds of RN for image inpainting, Basic RN (RN-B) and Learnable RN (RN-L). We finally introduce our image inpainting network using RN.

\subsection{Motivation for Region Normalization}
\subsubsection{Problem in Normalization.}
\begin{figure}[t]
	\begin{center}
		\includegraphics[width=.95\columnwidth]{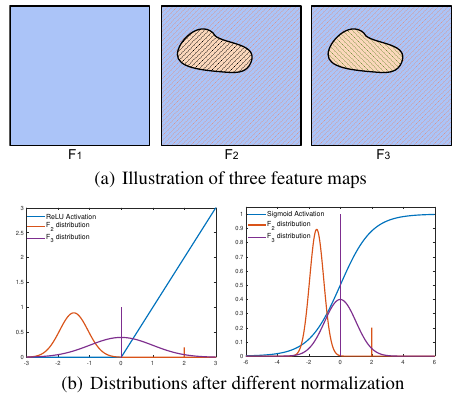}
	\end{center}
	
	\caption{(a) $\text{F}_1$ is the original feature map. $\text{F}_2$ with mask performs full-spatial normalization in all the regions. $\text{F}_3$ performs separate normalization in the masked and unmasked regions. (b) The distribution of $\text{F}_2$'s unmasked area has a shift to the nonlinear region, which easily causes the vanishing gradient problem. But $\text{F}_3$ does not have this problem.}
	\label{motivation}
\end{figure}
$\text{F}_1$, $\text{F}_2$ and $\text{F}_3$ are three feature maps of the same size, each with $n$ pixels, as shown in Figure \ref{motivation}. $\text{F}_1$ is the original uncorrupted feature map. $\text{F}_2$ and $\text{F}_3$ are the different normalization results of feature map with masked and unmasked areas. $n_m$ and $n_u$ are the pixel numbers of the masked and unmasked areas, respectively. Then $n = n_m + n_u$. Specifically, $\text{F}_2$ is normalized in all the areas. $\text{F}_3$ is normalized separately in the masked and unmasked areas. Assuming the masked region pixels have the max value $255$, the mean and standard deviation of three feature maps are listed as $\mu_1$, $\mu_2$, $\mu_{3m}$, $\mu_{3u}$, $\sigma_1$, $\sigma_{2}$, $\sigma_{3m}$ and $\sigma_{3u}$. The subscripts $1$ and $2$ represent the entire areas of $\text{F}_1$ and $\text{F}_2$, and $3m$ and $3u$ represent the masked and unmasked areas of $\text{F}_3$, respectively. The relationships are listed below:
\begin{equation}
{\mu_{3u}}={\mu_1}, {\sigma_{3u}={\sigma_1}}
\end{equation} 
\begin{equation}
{\mu_{3m}}=255, {\sigma_{3m}}=0
\end{equation}
\begin{equation}
{\mu_{2}}={\frac{n_u}{n}*\mu_{3u}+\frac{n_m}{n}*255}
\end{equation}
\begin{equation}
{{\sigma_{2}}^2}={\frac{n_u}{n}{\sigma_{3u}}^2+\frac{n_m*n_u}{n^2}(\mu_{3u}-255)^2}
\end{equation}

After normalizing the masked and unmasked areas together, $\text{F}_2$ unmasked area's mean has a shift toward $-255$ and its variance increases compared with $\text{F}_1$ and $\text{F}_3$. According to \cite{ioffe2015batch}, the normalization shifts and scales the distribution of features into a small region where the mean is zero and the variance is one. We take batch normalization (BN) as an example here.
For each point $x_i$
\begin{equation}
{x_i'}={\frac{x_i-\mu}{\sqrt{\sigma^2+\epsilon}}}
\end{equation}
\begin{equation}
{y_i}={\gamma x_i'+\beta}={BN_{\gamma,\beta}(x_i)}
\end{equation}
Compared with the $\text{F}_3$'s unmasked area, distribution of $\text{F}_2$'s unmasked area narrows down and shifts from $0$ toward $-255$.
Then, for both fully-connected and convolutional layer, the affine transformation is followed by an element-wise nonlinearity \cite{ioffe2015batch}:
\begin{equation}
{z}={g(BN(Wu))}
\end{equation}

Here $g(\cdot)$ is the nonlinear activation function such as ReLU or sigmoid. The BN transform is added immediately before the function, by normalizing $x = W u + b$. The $W$ and $b$ are learned parameters of the model.

As shown in Figure \ref{motivation}, in the ReLU and sigmoid activations, the distribution region of $\text{F}_2$ is narrowed down and shifted by the masked area, which adds the internal covariate shift and easily get stuck in the saturated regimes of nonlinearities (causing the vanishing gradient problem), wasting lots of time for $\gamma$, $\beta$ and W to fix the problem. However, $\text{F}_3$, normalized the masked and unmasked regions separately, reduces the internal covariate shift, which preserves the network capacity and improves training efficiency. 

Motivated by this, we design a spatial region-wise normalization named Region Normalization (RN).

\subsubsection{Formulation of Region Normalization.} 
Let $X \in \mathbb{R}^{N \times C \times H \times W}$ be the input feature. $N$, $C$, $H$ and $W$ are batch size, number of channels, height and width, respectively. Let $x_{n,c,h,w}$ be a pixel of $X$ and $X_{n,c} \in \mathbb{R}^{H \times W}$ be a channel of $X$ where ($n$, $c$, $h$, $w$) is an index along ($N$, $C$, $H$, $W$) axis. Given a region label map (mask) $M$, $X_{n,c}$ is divided into $K$ regions as follows:
\begin{equation}
X_{n,c}=R_{n,c}^{1} \cup R_{n,c}^{2} \cup ... \cup R_{n,c}^{K}
\end{equation} 

The mean and standard deviation of each region of a channel $R_{n,c}^{k}$ computed by:

\begin{equation}
\mu_{n,c}^{k}=\frac{1}{|R_{n,c}^{k}|}\sum_{x_{n,c,h,w} \in {R_{n,c}^{k}}}x_{n,c,h,w}
\end{equation} 
\begin{equation}
\sigma_{n,c}^{k}=\sqrt{\frac{1}{|R_{n,c}^{k}|}\sum_{x_{n,c,h,w} \in {R_{n,c}^{k}}}{(x_{n,c,h,w}-\mu_{n,c}^{k}})^2+\epsilon}
\end{equation} 
Here $k$ is a region index, ${|R_{n,c}^{k}|}$ is the number of pixels in region ${R_{n,c}^{k}}$ and $\epsilon$ is a small constant. The normalization of each region performs the following computation:
\begin{equation}
\hat R_{n,c}^{k} =\frac{1}{\sigma_{n,c}^{k}}(R_{n,c}^{k}-\mu_{n,c}^{k}) 
\end{equation} 

RN merges all normalized regions and obtains the region normalized feature as follows:
\begin{equation}
\hat X_{n,c}={\hat R_{n,c}^{1} \cup \hat R_{n,c}^{2} \cup ... \cup \hat R_{n,c}^{K}}
\end{equation} 

After normalization, each region is transformed separately with a set of learnable affine parameters $(\gamma_c^k, \beta_c^k)$.

\subsubsection{Analysis of Region Normalization.}
Here RN is an alternative to Instance Normalization (IN). RN degenerates into IN when region number $K$ equals to one. RN normalizes spatial regions on each channel separately as the spatial regions are not entirely dependent. We set $K=2$ for image inpainting in this work, as there are two obviously independent spatial regions in the input image: corrupted and uncorrupted regions. RN with $K=2$ is illustrated in Figure \ref{rn}. Note that RN is not limited to the IN style. Theoretically, RN can also be BN-style or based on other normalization methods.

\subsection{Basic Region Normalization} 
Basic RN (RN-B) normalizes and transforms corrupted and uncorrupted regions separately. This can solve the mean and variance shift problem of normalization and also avoid information mixing in affine transformation. RN-B is designed for using in early layers of the inpainting network, as the input feature has large corrupted regions, which causes severe mean and variance shifts.

Given an input feature $\bm{F} \in \mathbb{R}^{C \times H \times W}$ and a binary region mask $\bm{M} \in \mathbb{R}^{1 \times H \times W}$ indicating corrupted region, RN-B layer first separates each channel $\bm{F_c} \in \mathbb{R}^{1 \times H \times W}$ of input feature $\bm{F}$ into two regions $R_c^{1}$ ($e.g.$ uncorrupted region) and $R_c^{2}$ ($e.g.$ corrupted region) according to region mask $\bm{M}$. Let $x_{c,h,w}$ represent a pixel of $\bm{F_c}$ where $(c,h,w)$ is an index of $(C,H,W)$ axis.
The separation rule is as follow:
\begin{equation}
x_{c,h,w}\in \begin{cases}
R_c^{1} & \text{ if } \bm{M}(h,w)=1 \\ 
R_c^{2} & \text{ otherwise } 
\end{cases}
\end{equation}
RN-B then normalizes each region following Formula (9), (10) and (11) with region number $K=2$. Then we merge the two normalized regions $\hat R_c^{1}$ and $\hat R_c^{2}$ to obtain normalized channel $\bm{ \hat {F_c}}$. RN-B is a basic implement of RN and the region mask is obtained from the original inpainting mask. 

For each channel, there are two sets of learnable parameters $(\gamma_c^1, \beta_c^1)$ and $(\gamma_c^2, \beta_c^2)$ for affine transformation of each region. For ease of denotation, we denote $[\gamma_c^1, \gamma_c^2]$ as $\bm{\gamma}$, $[\beta_c^1, \beta_c^2]$ as $\bm{\beta}$. RN-B layer is showed in Figure \ref{2RNs}(a). 

\subsection{Learnable Region Normalization}
\begin{figure}[t]
	\begin{center}
		\includegraphics[width=.85\columnwidth]{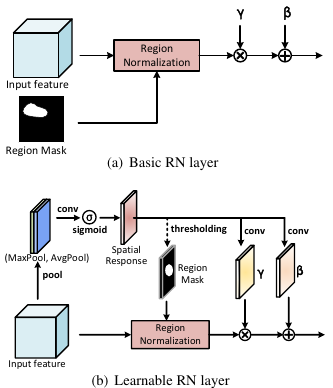}
	\end{center}
	\caption{Two kinds of RN: RN-B (a) and RN-L (b)}
	\label{2RNs}
\end{figure}
After passing through several convolutional layers, the corrupted regions are fused gradually and obtaining an accurate region mask from the original mask is hard. RN-L addresses the issue by automatically detecting corrupted regions and obtaining a region mask. To further improve the reconstruction, RN-L enhances the fusion of corrupted and uncorrupted regions by global affine transformation. RN-L boosts the corrupted region reconstruction in a soft way, which solves the mean and variance shift problem and also enhances the fusion. Therefore, RN-L is suitable for latter layers of the network. Note that, RN-L does not need a region mask and the affine parameters of RN-L are pixel-wise. RN-L is illustrated in Figure \ref{2RNs}(b).

RN-L generates a spatial response map by taking advantage of the spatial relationship of the features themselves. Specifically, RN-L first performs max-pooling and average-pooling along the channel axis. The two pooling operations are able to obtain an efficient feature descriptor \cite{zagoruyko2016paying,Woo_2018_ECCV}. RN-L then concatenates the two pooling results. RN-L is convolved on the two maps with sigmoid activation to get a spatial response map. The spatial response map is computed as:
\begin{equation}
\bm{M_{sr}} = \sigma(Conv([\bm{F_{max}},\bm{F_{avg}})]))
\end{equation} 
Here $\bm{F_{max}}\in \mathbb{R}^{1 \times H \times W}$ and $\bm{F_{avg}} \in \mathbb{R}^{1 \times H \times W}$ are the max-pooling and average-pooling results of the input feature $\bm{F} \in \mathbb{R}^{C \times H \times W}$. $Conv$ is the convolution operation and $\sigma$ is the sigmoid function. $\bm{M_{sr}}\in \mathbb{R}^{1 \times H \times W}$ is the spatial response map.
To get a region mask $\bm{M} \in \mathbb{R}^{1 \times H \times W}$ for RN, we set a threshold $t$ to the spatial response map:
\begin{equation}
\bm{M}(h,w) = \begin{cases}
1 & \text{ if } \bm{M_{sr}}(h,w) > t \\ 
0 & \text{ otherwise } 
\end{cases}
\end{equation}
We set threshold $t=0.8$ in this work. Note that the thresholding operation is only performed in the inference stage and the gradients do not pass through it during backpropagation.

Based on the mask $\bm{M}$, RN normalizes the input feature $\bm{F}$ and then performs a pixel-wise affine transformation. The affine parameters $\bm{\gamma} \in \mathbb{R}^{1 \times H \times W}$ and $\bm{\beta} \in \mathbb{R}^{1 \times H \times W}$ are obtained by convolution on the spatial response map $\bm{M_{sr}}$:
\begin{equation}
\bm{\gamma} = Conv(\bm{M_{sr}}), \bm{\beta} = Conv(\bm{M_{sr}})
\end{equation}

Note that the values of $\bm{\gamma}$ and $\bm{\beta}$ are expanded along the channel dimension in the affine transformation.
\begin{figure*}[t]
	\begin{center}
		\includegraphics[width=.885\linewidth]{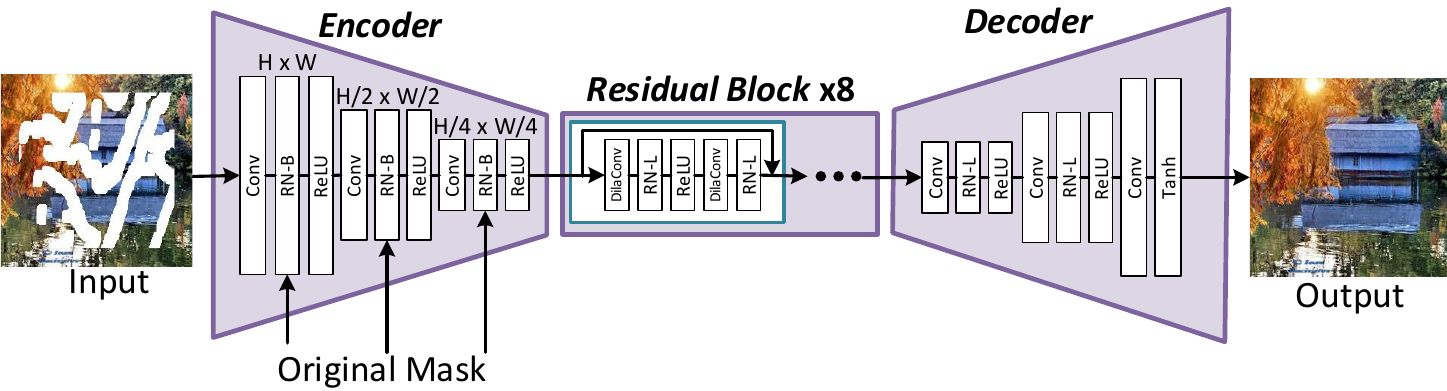}
	\end{center}
	\caption{Illustration of our inpainting model.}
	\label{arch}
\end{figure*}
The spatial response map $\bm{M_{sr}}$ has global spatial information. Convolution on it can learn a global representation, which boosts the fusion of corrupted and uncorrupted regions.

\subsection{Network Architecture}
EdgeConnect(EC) \cite{nazeri2019edgeconnect} consists of an edge generator and an image generator. The image generator is a simple yet effective network originally proposed by Johnson et al. \cite{johnson2016perceptual}. We use only the image generator as our backbone generator. We replace the original instance normalization (IN) of backbone generator to our two kinds of RN, RN-B and RN-L. Our generator architecture is shown in Figure \ref{arch}. Based the instruction of Section 3.2 and 3.3, we apply RN-B in the early layers (encoder) of our generator and RN-L in the intermediate and later layers (the residual blocks and decoder). Note that the input mask of RN-B is sampled from the original inpainting mask while RN-L does not need an external input as it generates region masks internally. We apply the same discriminators (PatchGAN \cite{isola2017image,zhu2017unpaired}) and loss functions (reconstruction loss, adversarial loss, perceptual loss and style loss) of the original backbone model to our model.

\section{Experiments}
We first compare our method with current state-of-the-art methods. We then conduct ablation study to explore the properties of RN and visualize our methods. Finally, we generalize RN to some other state-of-the-art methods.

\subsection{Experiment Setup}
We evaluate our methods on Places2 \cite{zhou2017places} and CelebA \cite{liu2015deep} datasets. We use two kinds of image masks: regular masks which are fixed square masks (occupying a quarter of the image) and irregular masks from \cite{liu2018image}. The irregular mask dataset contains 12000 irregular masks and the masked area in each mask occupies 0-60\% of the total image size. Besides, the irregular dataset is grouped into six intervals according to the mask area, $i.e.$0-10\%, 10-20\%, 20-30\%, 30-40\%, 40-50\% and 50-60\%. Each interval has 2000 masks.

\subsection{Comparison}
We compare our method to four current state-of-the-art methods and the baseline.

- CA: Contextual Attention \cite{yu2018generative}.

- PC: Partial Convolution \cite{liu2018image}.

- GC: Gated Convolution \cite{yu2019free}.

- EC: EdgeConnect \cite{nazeri2019edgeconnect}.

- Baseline: the backbone network we used. The baseline model use instance normalization instead of RN.
\subsubsection{Quantitative Comparisons}
We test all models on total validation data (36500 images) of Places2. We compare our model with CA, PC, GC, EC and the baseline. Three commonly used metrics are used: PSNR, SSIM \cite{wang2004image} with window size 11, and $l_1$ loss. We give the results of quantitative comparisons in Table \ref{Quantitative}. The second column is the area of irregular masks at testing time. Note that the $All$ in Table \ref{Quantitative} represents using all irregular masks (0-60\%) when testing. Our model surpasses all the comparing models on all three metrics. Compared to the baseline, our model improve PSNR by \textbf{0.73} dB and SSIM by \textbf{0.017}, and reduce $l_1$ loss (\%) by \textbf{0.25} in the $All$ case. 
\begin{table}[t]
	\centering
	\resizebox{.95\columnwidth}{!}{
		\begin{tabular}{c|c|cccc|cc}
			\hline
			\hline
			& Mask & CA & PC* & GC & EC & baseline & Ours \\
			\hline
			\multirow{4}{*}{PSNR$^{\uparrow}$}
			& 10-20\%     & 24.45 & 28.02 & 26.65          & 27.46 & 27.28 & \textbf{28.16} \\ 
			& 20-30\%     & 21.14 & 24.90 & 24.79          & 24.53 & 24.35 &\textbf{25.06} \\ 
			& 30-40\%     & 19.16 & 22.45 & \textbf{23.09} & 22.52 & 22.33 & 22.94          \\  
			& 40-50\%     & 17.81 & 20.86 & \textbf{21.72} & 20.90 & 20.96 & 21.21          \\  
			& All & 21.60 & 24.82 & 24.53          & 24.39 & 24.37 & \textbf{25.10} \\
			\hline
			\multirow{4}{*}{SSIM$^{\uparrow}$}
			& 10-20\%     & 0.891 & 0.869 & 0.882          & 0.920 & 0.914 &\textbf{0.926} \\
			& 20-30\%     & 0.811 & 0.777 & 0.836          & 0.859 & 0.851 &\textbf{0.868} \\
			& 30-40\%     & 0.729 & 0.685 & 0.782          & 0.794 & 0.784 &\textbf{0.804} \\ 
			& 40-50\%     & 0.651 & 0.589 & 0.721          & 0.723 & 0.711 &\textbf{0.734} \\ 
			& All & 0.767 & 0.724 & 0.807          & 0.814 & 0.806 & \textbf{0.823} \\
			\hline
			\multirow{4}{*}{$l_{1}$(\%)$^{\downarrow}$}
			& 10-20\%     & 1.81  & 1.14  & 3.01           & 1.58 & 1.24 & \textbf{1.10}  \\ 
			& 20-30\%     & 3.24  & 1.98  & 3.54           & 2.71 & 2.17 & \textbf{1.96}  \\ 
			& 30-40\%     & 4.81  & 3.02  & 4.25           & 3.93 & 3.19 & \textbf{2.90}  \\ 
			& 40-50\%     & 6.30  & 4.11  & 4.99           & 5.32 & 4.36 & \textbf{4.00}  \\ 
			& All & 4.21  & 2.80  & 3.79           & 2.83  &2.95& \textbf{2.70}  \\
			\hline
			
	\end{tabular}}
	\caption{Quantitative results on Places2 with models: CA \cite{yu2018generative}, PC \cite{liu2018image}, GC \cite{yu2019free}, EC \cite{nazeri2019edgeconnect}, the baseline, and ours(RN). All masks $i.e.$ masks with 0-60\% area. $^{\uparrow}$ higher is better. $^{\downarrow}$ lower is better. $^{*}$ the statistics are obtained from their paper.}
	\label{Quantitative}
\end{table}

\subsubsection{Qualitative Comparisons}
\begin{figure*}[t]
	\begin{center}
		\includegraphics[width=.95\linewidth]{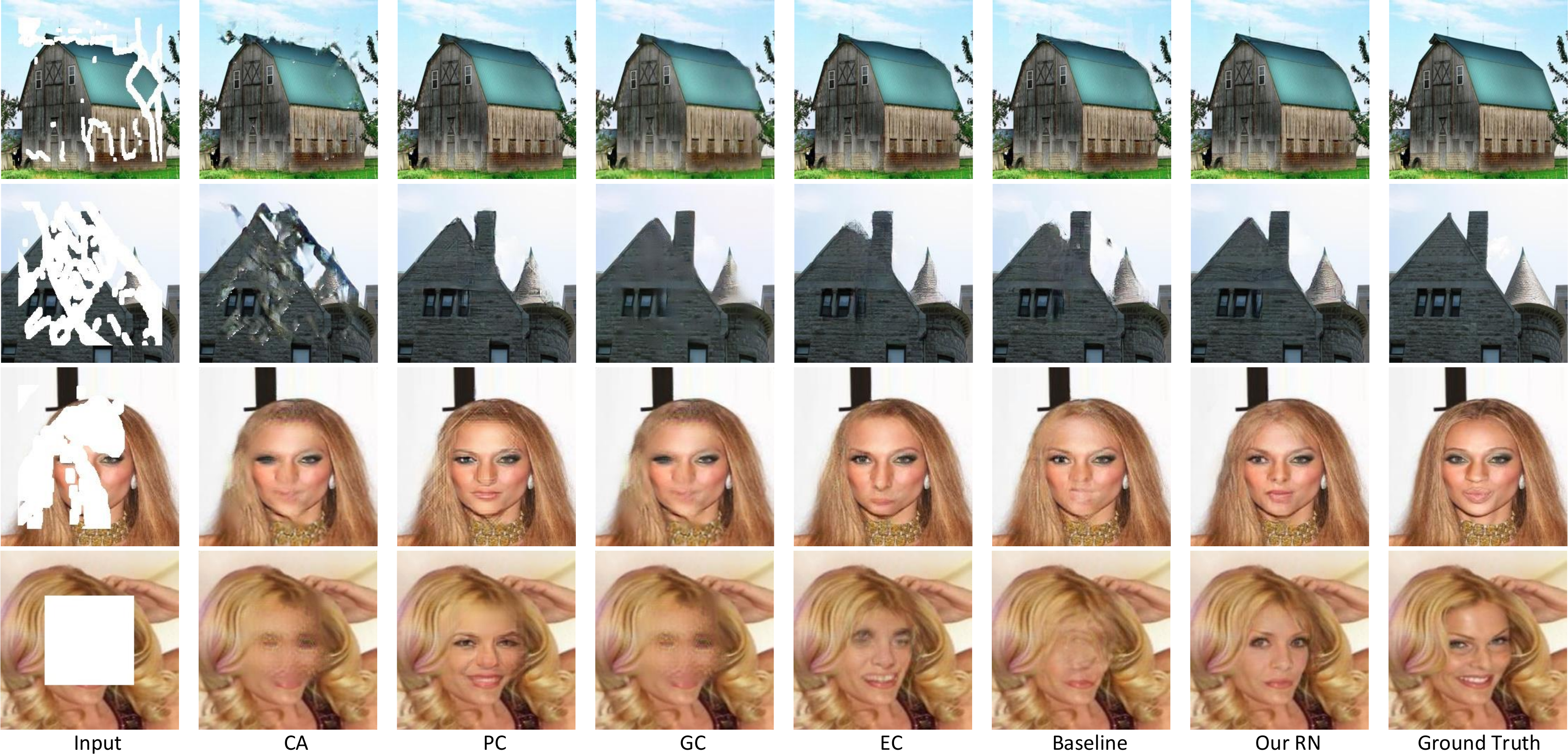}
	\end{center}
	\caption{Qualitative results with CA \cite{yu2018generative}, PC \cite{liu2018image}, GC \cite{yu2019free}, EC \cite{nazeri2019edgeconnect}, the baseline, and our RN. The first two rows are the testing results on Places2 and the last two are on CelebA.}
	\label{quality}
\end{figure*}

Figure \ref{quality} compares images generated by CA, PC, GC, EC, the baseline and ours. The first two rows of input images are taken from Places2 validation dataset and the last two rows are taken from CelebA validation dataset. In addition, the first three rows show the results in irregular mask case and the last row shows regular mask (fixed square mask in center) case. Our method achieves better subjective results, which benefits from RN-B's eliminating the impact of the mean and variance shifts on training, and RN-L's further boosting the reconstruction of corrupted regions.

\subsection{Ablation Study}
\begin{table}[t]
	\centering
	\resizebox{.95\columnwidth}{!}{
		\begin{tabular}{c|ccc|ccc}
			\hline
			\hline
			Arch. & Encoder & Res-blocks & Decoder &  PSNR & SSIM & $l_{1}$(\%) \\ \hline
			baseline & IN & IN   & IN  &   24.37  & 0.806   & 2.95   \\ 
			\textbf{1} & \textbf{RN-B} & \textbf{IN}   & \textbf{IN}   &   \textbf{24.88}  & \textbf{0.814}   & \textbf{2.77}  \\ 
			2 & RN-B   & RN-B  & IN   &  24.41  & 0.810 & 2.90  \\ 
			3 & RN-B   & RN-B  & RN-B & 24.59  & 0.812 & 2.85  \\ 
			4 & RN-B   & RN-L  & IN   & 25.02  & \textbf{0.823} & 2.71  \\  
			\textbf{5} & \textbf{RN-B}   & \textbf{RN-L}  & \textbf{RN-L} & \textbf{25.10}  & \textbf{0.823} & \textbf{2.70}  \\
			6 & RN-L   & RN-L  & RN-L & 24.53  & 0.812 & 2.86  \\ \hline
	\end{tabular}}
	\caption{ The influence of plugging location of RN-B and RN-L. The baseline uses inistance normalization (IN) in all three stages. The results are based on Places2.}
	\label{depth}
\end{table}

\begin{table}[t]
	\centering
	\resizebox{.55\columnwidth}{!}{
		\begin{tabular}{c|cccc}
			\hline
			\hline
			& None   & IN   	& BN   		& RN \\ \hline
			PSNR 		& 24.47  & 24.37  	& 24.24  	& \textbf{25.10}   \\  
			SSIM 		& 0.811  & 0.806  	& 0.806  	& \textbf{0.823}   \\ 
			$l_{1}$(\%) & 2.91	 & 2.95  	& 2.98  	& \textbf{2.70}  \\ \hline
	\end{tabular}}
	\caption{The final convergence results of different normalization methods on Places2. None means no normalization.}
	\label{finalconv}
\end{table}

\subsubsection{RN and Architecture}
\begin{figure}[ht]
	\begin{center}
		\includegraphics[width=0.95\linewidth]{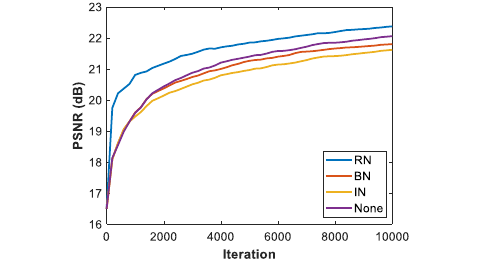}
	\end{center}
	\caption{The PSNR results of different normalization methods in the first 10000 iterations on Places2. None means no normalization.}
	\label{psnr_training}
\end{figure}
We first explore the source of gain for our methods and the best strategy to apply two kinds of RN: RN-B and RN-L. We conduct ablation experiments on the backbone generator, which has three stages: an encoder, followed by eight residual blocks and a decoder. We plug RN-B and RN-L in different stages and obtain six architectures (Arch.1-6) as shown in Table \ref{depth}. The results in Table \ref{depth} show the effectiveness of our use of RN: apply RN-B in the early layers (encoder) to solve the mean and variance shifts caused by large-area uncorrupted regions; apply RN-L in the later layers to solve the the mean and variance shifts and boost the fusion of two kinds of regions. Arch.1 only applies RN-B in the encoder and achieves a significant performance boost, which directly shows the RN-B's effectiveness. Arch.2 and 3 reduce the performance as RN-B can hardly obtain an accurate region mask in the latter layers of the network after passing through several convolutional layers. Arch.4 is beyond Arch.1 by adding RN-L in the middle residual blocks. Arch.5 (Our method) further improves the performance of Arch.4 by applying RN-L in both the residual blocks and the decoder. Note that Arch.6 uses RN-L to the encoder and its performance is reduced compared to Arch.5 ,since RN-L, a module of soft fusion, unavoidably mixing up information from corrupted and uncorrupted regions and washing away information from the uncorrupted regions. The above results verify the effectiveness of our use of RN-B and RN-L that we explain in Section 3.2 and 3.3.

\subsubsection{Comparisons with Other Normalization Methods}
To verify our RN is more effective in training of the inpainting model, we compare our RN with a none-normalization method and two full-spatial normalization methods, batch normalization (BN) and instance normalization (IN), based on the same backbone. We show the PSNR curves in the first 10000 iterations in Figure \ref{psnr_training} and the final convergence results (about 225,000 iterations) in Table \ref{finalconv}. The experiments are on Places2. Note that no normalization (None) is better than full-spatial normalization (IN and BN), and RN is better than no normalization by eliminating the mean and variance shifts and taking advantage of normalization technique at the same time.

\subsubsection{Threshold of Learnable RN}
\begin{figure}[t]
	\begin{center}
		\includegraphics[width=.95\linewidth]{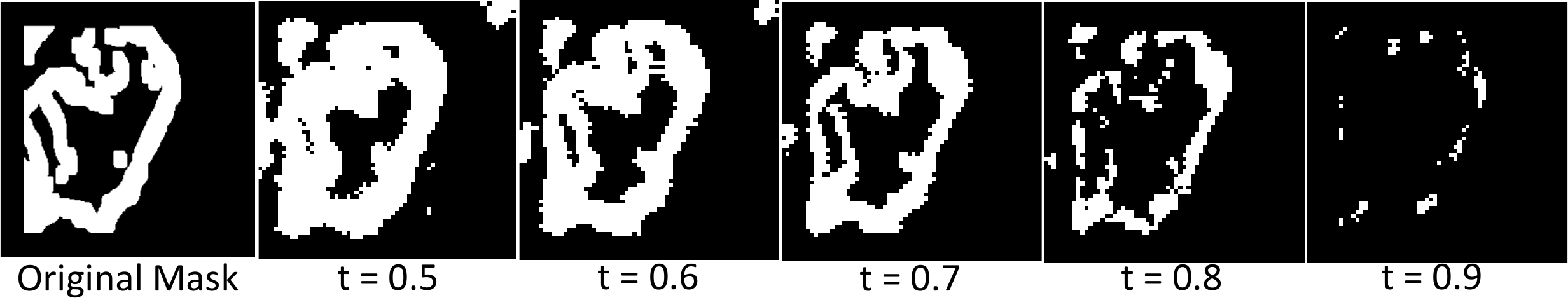}
	\end{center}
	\caption{The generated mask with different threshold $t$ of the first RN-L layer in the sixth residual block.}
	\label{mask_change_figure}
\end{figure}

Threshold $t$ is set in Learnable RN to generate a region mask from the spatial response map. The threshold affects the accuracy of the region mask and further affects the power of RN. We conduct a set of experiments to explore the best threshold. The PSNR results on Places2 and CelebA show that RN-L achieves the best results when threshold $t$ equals to 0.8, as shown in Table \ref{threshold}. We show the generated mask of the first RN-L layer in the sixth residual block ($R6RN1$) as an example in Figure \ref{mask_change_figure}. The generated mask of $t=0.8$ is likely to be the most accurate mask in this layer.
\begin{table}[t]
	\centering
	\resizebox{.76\columnwidth}{!}{
		\begin{tabular}{c|ccccc}
			\hline
			\hline
			$t$     & 0.5   & 0.6   & 0.7   & 0.8            & 0.9   \\ \hline
			Places2 & 23.85 & 24.90 & 24.96 & \textbf{25.10} & 24.93 \\ 
			CelebA  & 27.36 & 27.92 & 28.45 & \textbf{28.51} & 23.73 \\ \hline 
	\end{tabular}}
	\caption{The PSNR results with different threshold $t$ on Places2 and CelebA datasets.}
	\label{threshold}
\end{table}

\subsubsection{RN and Masked Area}
We explore the mask area's influence to RN. Based the theoretical analysis in Section 3.1, the mean and variance shifts become more severe as mask area increases. Our experiments on CelebA show that the advantage of our RN becomes more significant as the mask area increases, as shown in Table \ref{maskarea}. We use $l_1$ loss to evaluate the results.
\begin{table}[t]
	\centering
	\resizebox{.95\columnwidth}{!}{
		\begin{tabular}{c|cccccc}
			\hline
			\hline
			Mask     & 0-10\%   & 10-20\%   & 20-30\%   & 30-40\%  & 40-50\% & 50-60\%   \\ \hline
			baseline & 0.26 & 0.69 & 1.28 & 2.02 & 2.92 & 4.83 \\ 
			RN  & \textbf{0.23} & \textbf{0.62} & \textbf{1.18} & \textbf{1.85} & \textbf{2.68} & \textbf{4.52} \\ \hline
			Change & -0.03 & -0.07 & -0.10 & -0.17 & -0.24 & -0.31 \\ \hline 
	\end{tabular}}
	\caption{The testing $l_1$(\%) loss with different mask area on CelebA. RN's advantage becomes more significant as the mask area increases.}
	\label{maskarea}
\end{table}

\begin{table}[t]
	\centering
	\resizebox{.85\columnwidth}{!}{
		\begin{tabular}{l|ll|ll|ll}
			\hline
			\hline
			& CA & RN-CA & PC & RN-PC & GC & RN-GC \\ 
			\hline
			PSNR
			&  21.60 & \textbf{24.12} &  24.82  & \textbf{25.32} & 24.53 & \textbf{24.55} \\
			
			SSIM
			&  0.767 & \textbf{0.842} &  0.724  & \textbf{0.829} & \textbf{0.807} & \textbf{0.807} \\
			
			$l_{1}(\%)$
			&  4.21 & \textbf{3.17}   &  2.80   & \textbf{2.61}  & 3.79  & \textbf{3.75}  \\ 
			\hline
			
	\end{tabular}}
	\caption{The results of applying RN to different backbone networks: CA \cite{yu2018generative}, PC \cite{liu2018image} and GC \cite{yu2019free}. The results is based on Places2.}
	\label{generalize}
\end{table}

\begin{figure}[ht!]
	\begin{center}
		\includegraphics[width=.95\linewidth]{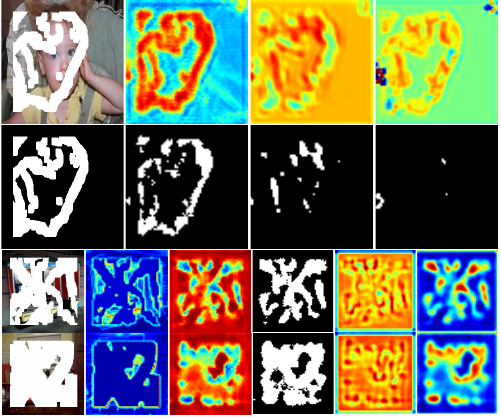}
	\end{center}
	\caption{Visualization of our method. The top two rows are illustrated the changes of the spatial response and generated mask in different locations of the network: the first RN-L in the sixth residual block, the second RN-L in the seventh residual block and the second RN-L in the eighth residual block. In the last two rows, from left to right: input, encoder result, spatial response map, generated mask, gamma map and beta map of the first RN-L in the seventh residual block.}
	\label{visual}
\end{figure}

\subsection{Visualization}
We visualize some features of the inpainting network to verify our method. We show the changes of the spatial response and generated mask of RN-L as the network deepens in the top two rows of Figure \ref{visual}. The mask changes in different layers as the fusion effect of passing through convolutional layers. RN-L can detect potentially corrupted regions consistently. From the last two rows of Figure \ref{visual} we can see: (1) the uncorrupted regions in the encoded feature are well preserved by using RN-B; (2) RN-L can distinguish between potentially different regions and generate a region mask; (3) gamma and beta maps in RN-L perform a pixel-level transform on potentially corrupted and uncorrupted regions distinctively to help the fusion of them.

\subsection{Generalization Experiments}
RN-B and RN-L are plug-and-play modules in image inpainting networks. We generalize our RN (RN-B and RN-L) to some other backbone networks: CA, PC and GC. We apply RN-B to their early layers (encoder) and RN-L to the later layers. CA and GC are two-stage (coarse-to-fine) inpainting networks and the coarse result is the input of the refinement network. The corrupted and uncorrupted regions of the coarse result is typically not particularly obvious, thus we only apply RN to the coarse inpainting networks of CA and GC. The results on Places2 are shown in Table \ref{generalize}. The RN-applied CA and PC achieve a significant performance boost by \textbf{2.52} and \textbf{0.5} dB PSNR respectively. The gain on GC is not very impressive. A possible reason is that gated convolution of GC greatly smoothes features which make RN-L hard to track potentially corrupted regions. Besides, GC's results are typically blurry as shown in Figure \ref{quality}.

\section{Conclusion}
\begin{figure}[t]
	\begin{center}
		\includegraphics[width=.95\linewidth]{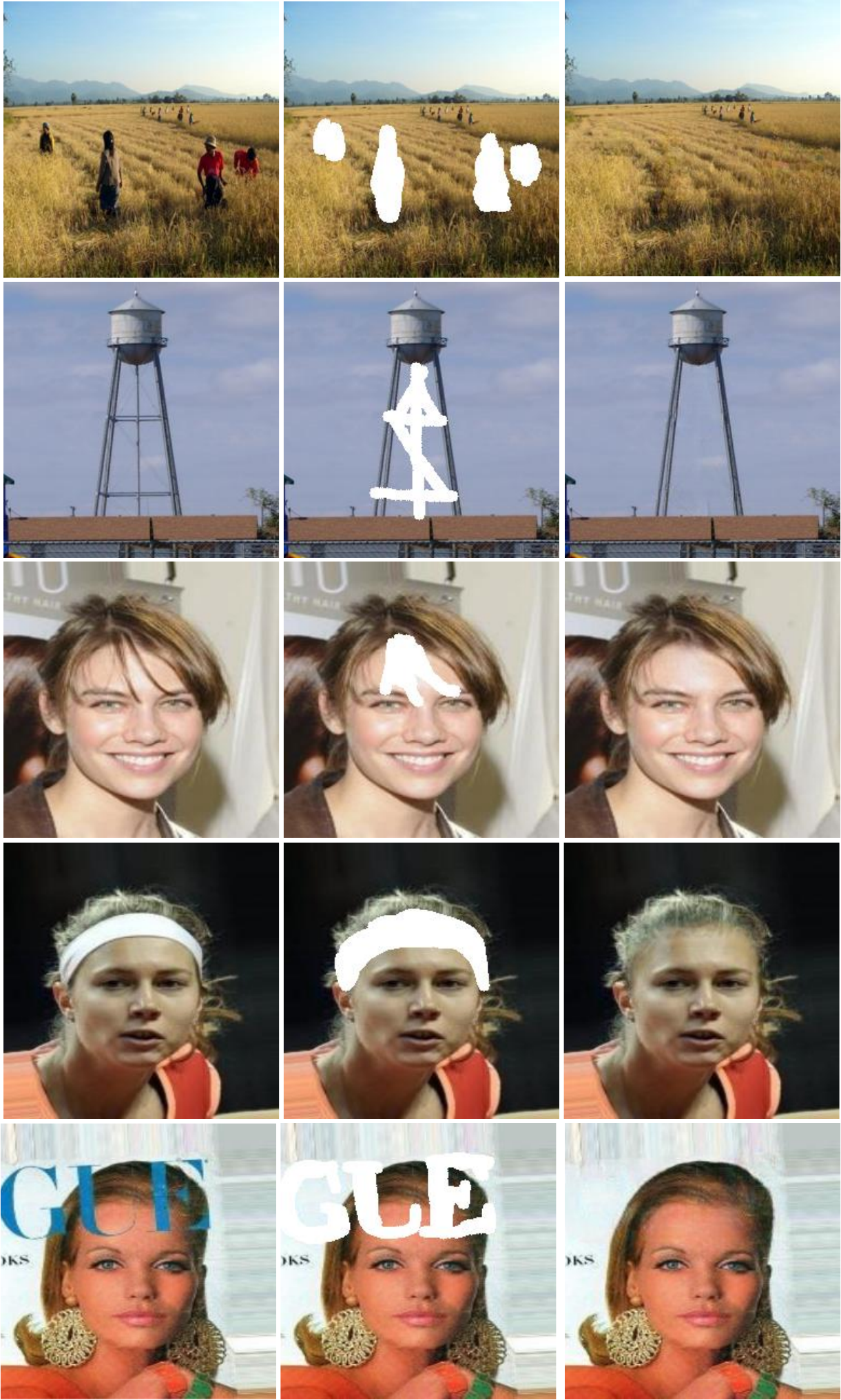}
	\end{center}
	\caption{Our results in real use cases.}
	\label{real}
\end{figure}

In this work, we investigate the impact of normalization on inpainting network and show that Region Normalization (RN) is more effective for image inpainting network, compared with existing full-spatial normalization. The proposed two kinds of RN are plug-and-play modules, which can be applied to other image inpainting networks conveniently. Additionally, our inpainting model works well in real use cases such as object removal, face editing and image restoration, as shown in Figure \ref{real}. 

In the future, we will explore RN for other supervised vision tasks such as classification, detection and so on. 

\bibliography{833_reference}
\bibliographystyle{aaai}

\end{document}